\documentclass{article}

\usepackage[final]{corl_2026} 
\usepackage{graphicx}
\usepackage{booktabs}
\usepackage{booktabs}
\usepackage{makecell}
\usepackage{booktabs}
\usepackage{makecell}
\usepackage{graphicx}
\usepackage{wrapfig}
\usepackage{caption}
\usepackage{amsmath}
\usepackage{amssymb}
\usepackage{booktabs}
\usepackage{multirow}
\usepackage{array}
\usepackage{amsmath}
\usepackage{amssymb}
\usepackage{algorithm}
\usepackage{algorithmic}
\usepackage{booktabs}
\usepackage{tabularx}
\usepackage{array}
\definecolor{deepred}{RGB}{139,0,0}
\usepackage[table]
{xcolor}
\definecolor{darkred}{RGB}{150,0,0}

\title{FocusVLA: Future-Intent Guided Dynamic View Routing for Vision-Language-Action Models}

\author{
Huixin Xu$^{1,2,3,4}$ \quad
Ziya Yu$^{1,2,3,5}$ \quad
Peng Li$^{2,3,1,*}$\\
$^{1}$ Nanjing Institute of Software Technology \\
$^{2}$ Institute of Software, Chinese Academy of Sciences\\
$^{3}$ University of Chinese Academy of Sciences, Nanjing\\
$^{4}$ Nanjing University of Posts and Telecommunications \quad $^{5}$ Hohai University\\
$^{*}$ Corresponding author\\ 
}

\begin{document}
\maketitle

\begin{figure}[h]
    \centering
    \vspace{-3em}
    \includegraphics[width=\linewidth]{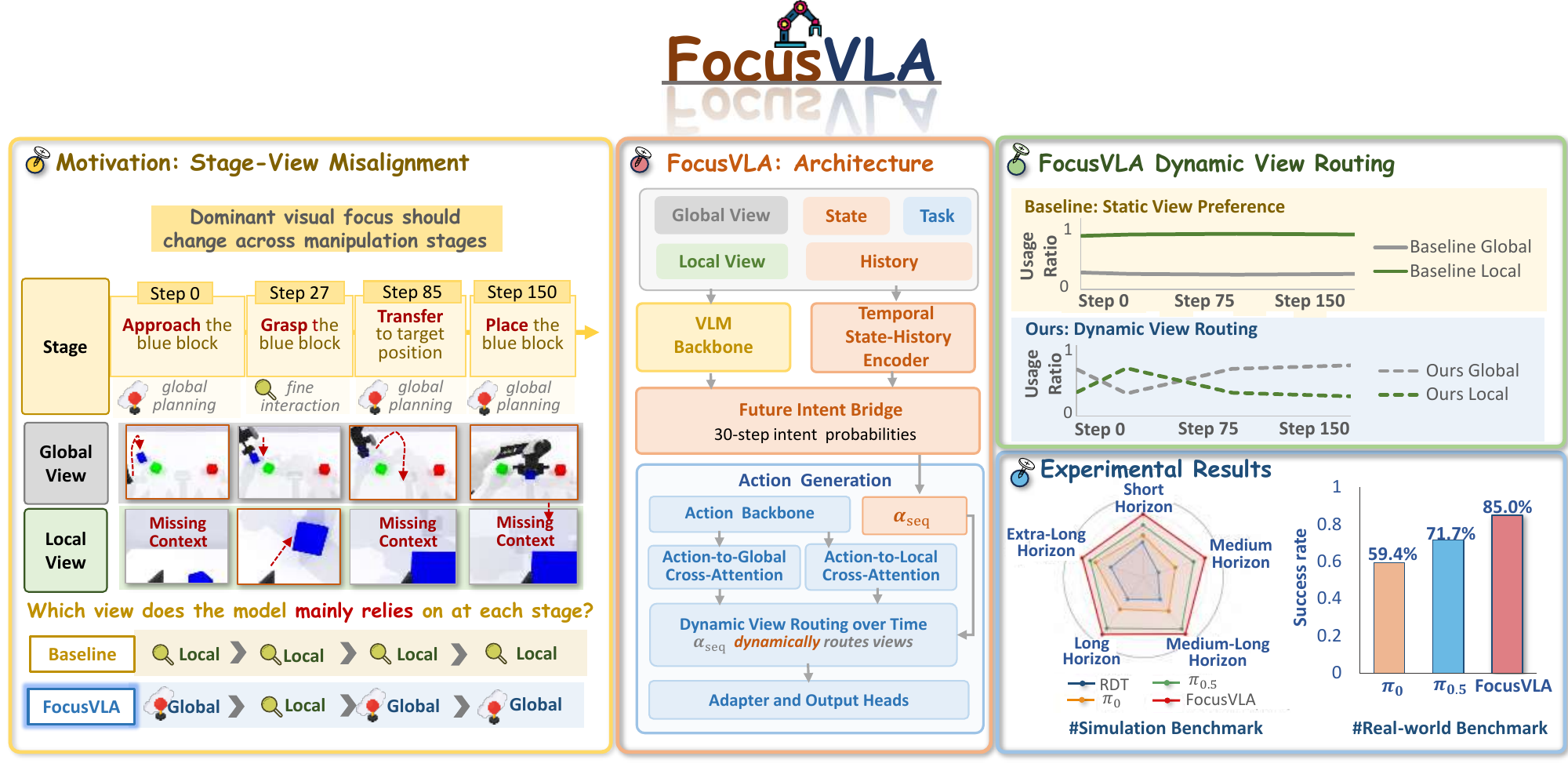}
    \caption{
    Overview of FocusVLA. Static multi-view fusion can cause view-dependency inertia, making the policy rely on a fixed view across task stages. FocusVLA predicts future action intent from multimodal observations and action history, and uses it to guide dynamic view routing during action generation, improving multi-view utilization in both simulation and real-world tasks.
    }
    \label{fig:overview}
\end{figure}

\begin{abstract}
    Vision-language-action (VLA) models have become an important paradigm for robotic manipulation by unifying visual perception, language understanding, and action generation. Since reliable manipulation requires sufficient observation of objects, end-effectors, and their spatial relationships, multi-view inputs are widely used to provide complementary spatial information. However, our experiments show that existing VLA models do not fully exploit such multi-view complementarity. The policy tends to persistently rely on the wrist view, lacking the ability to dynamically adjust global-local view dependency across manipulation stages, which limits stage adaptability and robustness. To address this issue, we propose FocusVLA, a stage-aware multi-view routing framework. FocusVLA predicts future interaction intent from historical actions and uses it to guide dynamic view routing during action generation. By adaptively coordinating global and local observations across different task stages, FocusVLA improves multi-view utilization. On RoboTwin 2.0, FocusVLA improves the average success rate over the strongest baseline $\pi_{0.5}$ by 10.10 pp. In real-world evaluation, FocusVLA improves the success rate from 71.7\% to 85.0\%, achieving a 13.3\% gain over $\pi_{0.5}$. Moreover, the intent prediction and dynamic routing modules of FocusVLA can be integrated into existing VLA frameworks without modifying their core architectures.
\end{abstract}



\section{Introduction}
	Recent Vision-Language-Action (VLA) models~\citep{kim2025fine, intelligence2025pi_, black2024pi_0, liu2025rdt, bjorck2025gr00t, wu2026pragmatic, kim2024openvla, yang2025instructvla, fan2025long, qu2025spatialvla, zhang2025ta} have shown strong potential for generalist robotic manipulation by unifying visual perception, language understanding, and action generation. In real-world manipulation, robust action generation often requires visual observations that can handle occlusions, viewpoint variations, and fine-grained interactions. Therefore, multi-view inputs are widely used to provide complementary visual information~\citep{seo2023multi, lan2025bfa, jangir2022look}: third-person cameras provide global scene context for target localization and motion planning, while wrist cameras provide local cues for alignment, grasping, and contact-rich execution. However, existing VLA models typically fuse visual observations with language instructions and robot states into a fixed context for action generation. Although this paradigm effectively integrates multimodal information, it rarely considers that the policy's visual dependency may change across manipulation stages. 
    
    Our empirical analysis further shows that, under the static fusion paradigm, the model is prone to view-dependency inertia during action generation: the policy may develop a strong reliance on a particular view at an early task stage according to the current manipulation demand. Although such reliance may be reasonable at the early stage, the lack of explicit awareness of task progress causes the early-formed observation preference to persist into later stages, making it difficult for the policy to dynamically adjust the contributions of different views to action generation according to changing manipulation requirements. This phenomenon suggests that improving multi-view utilization requires not only fusing multi-view information, but also enabling the policy to dynamically select and coordinate different view observations during action execution.  Motivated by these observations, we aim to mitigate the stage-view misalignment caused by static fusion. Achieving this goal requires answering two key questions: (1) how can a VLA policy acquire a representation that reflects task progress and use it to infer the visual requirements of upcoming manipulation stages? (2) how can such a representation be transformed into effective multi-view routing signals without disrupting the pretrained VLA backbone? 
    
    To answer these questions, we propose \textbf{FocusVLA}, a stage-aware dynamic view routing framework for multi-view VLA policies, as shown in Fig.~\ref{fig:overview}. FocusVLA predicts future action intent from multimodal observations and short-term action history, and uses it to generate action-step-wise routing weights that dynamically coordinate visual features from different views during action generation. The framework is non-invasive, requiring no modification to the visual encoder, language model, or action-generation backbone. We evaluate FocusVLA in both simulation and real-world settings. The simulation experiments cover all 50 RoboTwin~\citep{chen2025robotwin} tasks, and the real-world experiments include 6 representative manipulation tasks with diverse object configurations, viewpoint variations, and interaction requirements. Results show that FocusVLA consistently outperforms static multi-view fusion baselines, demonstrating its effectiveness in mitigating stage-view misalignment and improving policy adaptability. 
    
    In summary, our contributions are: (1) We identify and quantify a temporal failure in static multi-view VLA policies: view dependence formed at earlier stages persists after visual requirements change, causing stage--view misalignment. (2) We propose FocusVLA, a task-progress-aware visual modulation framework that converts future action intent into action-step-wise view-routing weights for non-invasive dynamic multi-view modulation. (3) We evaluate FocusVLA on all 50 RoboTwin simulation tasks and 6 real-world manipulation tasks, demonstrating improved adaptability and robustness over static multi-view baselines.


\section{Related Works}
\subsection{Vision-Language-Action Models.}
\begin{wrapfigure}[9]{r}{0.45\textwidth}
    \centering
    \vspace{-1em} \includegraphics[width=0.45\textwidth]{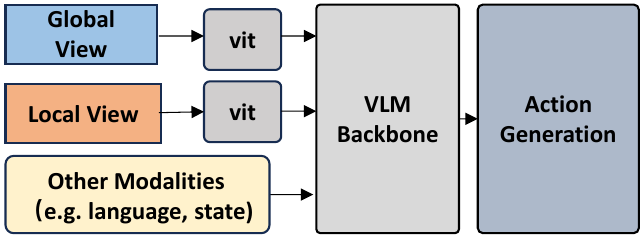}
    \caption{
    Fusion pathway of multi-view perception in VLA models.
    }
    \label{fig:normal_arc}
\end{wrapfigure}

Vision-Language-Action (VLA) models unify visual perception, language understanding, and action generation for robotic control~\citep{brohan2022rt, zitkovich2023rt, team2024octo, wu2024unleashing, cheang2024gr, wen2025dexvla, shukor2025smolvla, team2025gemini, zhao2025cot, zhang2026dreamvla, lu2025vla}. 
Recent models have improved generalist manipulation through diverse action generation paradigms, model scaling, and data scaling~\citep{o2024open, khazatsky2024droid}.
However, most existing VLA models still follow a relatively fixed perception-to-action pipeline. 
As shown in Fig.~\ref{fig:normal_arc}, visual observations are first encoded into image tokens or visual features, fused with language instructions and robot states, and then provided as context to the action generation module~\citep{intelligence2025pi_, black2024pi_0, kim2024openvla, zitkovich2023rt}. 
While this paradigm effectively integrates multimodal information, it rarely considers that the policy's visual dependency may change across different manipulation stages. 
This limitation becomes especially important in multi-view settings, where global and local views provide complementary but stage-dependent information~\citep{seo2023multi, lan2025bfa, jangir2022look, guhur2023instruction, goyal2023rvt, shridhar2023perceiver}.
Unlike prior works that mainly improve action decoders or scale VLA backbones, FocusVLA introduces a lightweight, architecture-agnostic view-routing module on the action-generation side. By predicting future action intent, FocusVLA dynamically modulates global and local visual contexts, enabling stage-aware multi-view adaptation without modifying existing VLA architectures.

\subsection{Multi-View Perception and Dynamic Visual Routing.}
Multi-view perception provides complementary information for robotic manipulation~\citep{seo2023multi, jangir2022look}: third-person views capture global scene context, while wrist views provide fine-grained hand--object interaction cues. Existing methods exploit such observations through representation learning, 3D reconstruction, feature aggregation, cross-attention, or perspective-adaptive fusion~\citep{gervet2023act3d, acar2023visual, lin2021multi, akinola2020learning, chen2025temporal, mao2025omnid, ke20243d}; recent VLA-oriented works further demonstrate the value of multi-view inputs for action generation~\citep{bian2025vla, fan2026peafowl, li2025uni, sarowar2026gst}. Most methods nevertheless adopt fixed view organization or generic fusion. BFA~\citep{lan2025bfa} addresses redundancy from equally treating current multi-view features by predicting vision-based view weights and fusing the features before an imitation policy. FocusVLA instead addresses temporal mismatch between learned view dependence and changing future action requirements in chunked VLA. The two methods differ in four key aspects: (1) prediction horizon: BFA produces one current-view allocation, whereas FocusVLA predicts $H$ step-specific future routes from multimodal context; (2) fusion site: BFA fuses views before the policy, whereas FocusVLA performs per-step fusion within action generation; (3) label construction: BFA uses offline VLM annotations and task-specific mappings, whereas FocusVLA derives task-agnostic labels from trajectories; and (4) intent token: FocusVLA introduces chunk-level intent conditioning absent from BFA. Together, these designs enable future-intent-guided coordination of global and local visual contexts without altering the core VLA backbone.


\section{Method}
\subsection{Motivation: Diagnosing Stage-Invariant View Dependency in Multi-View VLA}

\begin{wrapfigure}[14]{r}{0.5\textwidth}
\vspace{-1.5em} 
    \centering
    \includegraphics[width=0.5\textwidth]{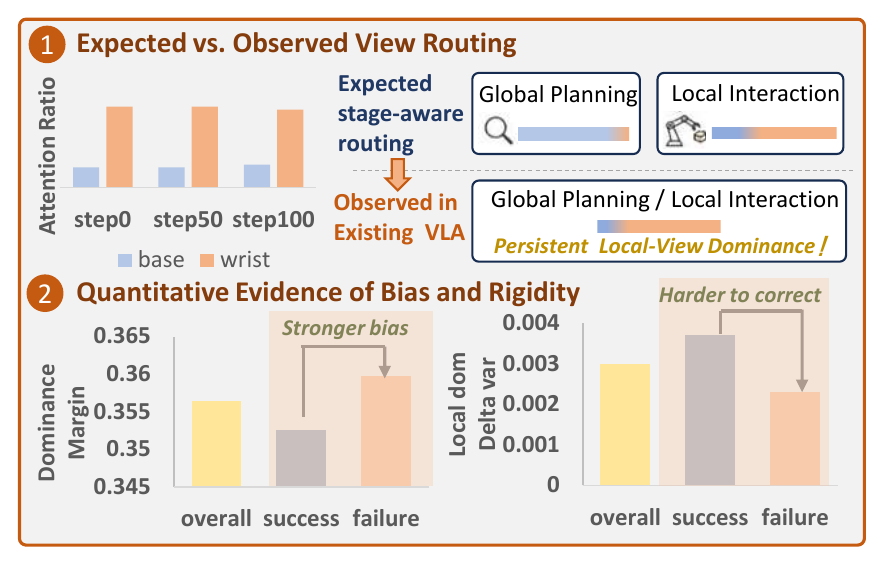}
    \caption{
    Motivation analysis of stage-insensitive view dependency in multi-view VLA models.
    }
    \label{fig:focusvla_analysis}
\end{wrapfigure}
To diagnose view usage in existing multi-view VLA models, we analyze the $\pi_{0.5}$~\citep{intelligence2025pi_} multi-view baseline by computing per-step attention ratios over global and local visual contexts. 
At action step \(t\), \(a_t^{\mathrm{global}}\) and \(a_t^{\mathrm{local}}\) denote the normalized attention ratios for the global view and aggregated local views, respectively, with \(a_t^{\mathrm{global}} + a_t^{\mathrm{local}} = 1\). 
When multiple local cameras are used, their attention ratios are summed as \(a_t^{\mathrm{local}}\), allowing a unified comparison between global and local reliance.
We quantify view usage with two metrics. 
The first is the \textit{Dominance Margin}, which measures global-local imbalance:
\begin{equation}
    D_t = \left| a_t^{\mathrm{global}} - a_t^{\mathrm{local}} \right|,
    \qquad
    D = \frac{1}{T}\sum_{t=1}^{T} D_t .
\end{equation}
The second is the temporal variation of local-view reliance:
\begin{equation}
    L_t = a_t^{\mathrm{local}},
    \qquad
    \Delta L_t = L_t - L_{t-1},
    \qquad
    V_{\Delta L} = \mathrm{Var}\left(\{\Delta L_t\}_{t=2}^{T}\right).
\end{equation}
A larger \(D\) indicates stronger dominance of one visual context, while a smaller \(V_{\Delta L}\) suggests weaker adjustment of visual reliance over time.
Empirically, both successful and failed rollouts exhibit imbalanced global-local reliance, typically biased toward the local visual context. This suggests that local-view preference itself is common and not necessarily harmful. However, failed rollouts show a larger Dominance Margin and a smaller \(V_{\Delta L}\), indicating stronger and less adaptive local-view dependency over time.

Since attention ratios do not necessarily imply causal reliance~\citep{jain2019attention, serrano2019attention, chefer2021transformer, abnar2020quantifying}, we further conduct a 
\begin{wraptable}[6]{r}{0.45\textwidth}
    \centering
    \vspace{-1em}
    \caption{Stage-conditioned View Masking.}
    \label{tab:view_masking}
    \scriptsize
    \setlength{\tabcolsep}{3pt}
    \begin{tabular*}{0.45\textwidth}{@{\extracolsep{\fill}}llcc@{}}
    \toprule[1.0pt]
    \textbf{Model} & \textbf{Masking Mode} & \textbf{SR} & \textbf{$\Delta$SR} \\
    \midrule
    Baseline & No Mask & 85.0\% & -- \\
    Baseline & Mask \textbf{\textcolor{darkred}{Global}}@Planning 
    & 65.0\% & \textcolor{darkred}{-20.0 pp} \\
    Baseline & Mask \textbf{\textcolor{darkred}{Local}}@Planning  
    & 20.0\% & \textcolor{darkred}{-65.0 pp} \\
    \bottomrule[1.0pt]
    \end{tabular*}
\end{wraptable}
stage-conditioned view masking experiment on the \textit{Stack Blocks Three} task, using the $\pi_{0.5}$~\citep{intelligence2025pi_} multi-view baseline. 
For each masking setting, we evaluate 20 rollouts. 
During global planning stages, we separately mask the global view and the local view. If the policy effectively uses global observations for planning, masking the global view should cause a larger performance drop.
However, Table~\ref{tab:view_masking} shows the opposite pattern: masking the global view reduces the success rate from 85.0\% to 65.0\%, whereas masking the local view causes a much larger drop to 20.0\%. This suggests that the static multi-view baseline remains highly dependent on local observations even during global-planning stages.

Together, the attention analysis and masking intervention provide complementary evidence for stage-insensitive view dependency. The attention metrics indicate that failed rollouts exhibit stronger local-view dominance and weaker temporal adjustment of visual reliance, while the intervention results further show that the baseline fails to shift its reliance toward the global view when global scene context is needed. This limitation motivates FocusVLA, which uses future intent to dynamically regulate global and local visual contexts during action generation.

\subsection{The FocusVLA Model Architecture}

\begin{figure}[t]
    \centering
    \includegraphics[width=0.95\linewidth]{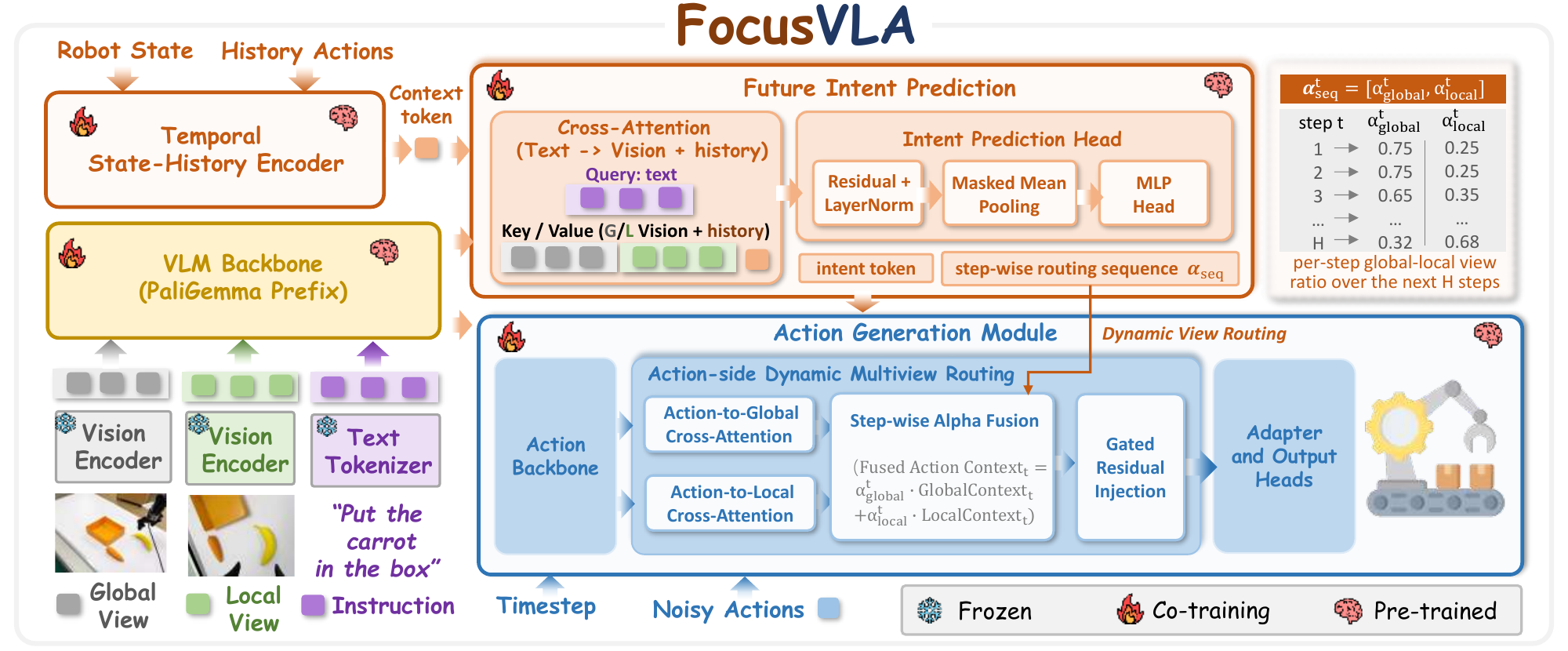}
    \caption{
    Architecture of FocusVLA. Given multi-view observations, language instructions, robot states, and action history, FocusVLA predicts future action intent and uses it to guide action-side dynamic view routing for adaptive multi-view feature modulation.
    }
    \label{fig:methods}
    \vspace{-1.5em}
\end{figure}

As shown in Fig.~\ref{fig:methods}, \textbf{FocusVLA} is built upon the $\pi_{0.5}$~\citep{intelligence2025pi_} multi-view VLA backbone. Given global and local observations \(o_t=\{I_t^g, I_t^l\}\), a language instruction \(L\), robot state \(s_t\), and short-term action history \(h_t\), FocusVLA predicts a future action chunk \(A_t=[a_t,a_{t+1},\dots,a_{t+H-1}]\).
Different from static multi-view fusion, FocusVLA uses future action trends to guide view dependency, allowing visual reliance to evolve with task progress. It consists of two modules: \textbf{Future Intent Bridge}, which predicts future interaction intent and produces a chunk-level intent token with step-wise routing weights \(\alpha_{1:H}\), and \textbf{Action-Side Dynamic Routing}, which uses these signals to regulate the contributions of global and local visual contexts during action generation. This design enables stage-aware multi-view modulation without modifying the pretrained VLA backbone.

\subsubsection{Future Intent Bridge}

Future Intent Bridge aims to infer future interaction trends from the current multimodal context. Given the vision-language features from the VLA prefix backbone, together with the robot state and short-term action history, the module first encodes recent action dynamics into a temporal summary. It then performs cross-modal attention, using language features as queries and visual/state-history features as keys and values, to obtain a contextualized representation for future intent prediction.

Based on this representation, the bridge predicts a step-wise intent sequence over the future action horizon, 
\(I_t = [i_t^{(1)}, i_t^{(2)}, \dots, i_t^{(H)}]\), 
where \(i_t^{(k)} \in [0,1]\) denotes the probability that the \(k\)-th future action step requires local interaction-oriented visual evidence. Accordingly, \(1-i_t^{(k)}\) can be interpreted as the probability of workspace-level motion, such as target localization or object transport. The predicted intent sequence is used in two ways. First, it is aggregated into a chunk-level intent token, 
\(z_{\mathrm{intent}}=\mathrm{MLP}_{\mathrm{token}}(\mathrm{Pool}(I_t))\), 
which summarizes the overall future interaction trend. Second, each step-wise intent prediction is converted into global--local routing weights through a routing function,
\(\alpha^{(k)}=f_{\mathrm{route}}(i_t^{(k)})\),
where \(\alpha^{(k)}=[\alpha_g^{(k)}, \alpha_l^{(k)}]\) consists of nonnegative weights that sum to one and controls the relative contributions of global and local visual contexts for the \(k\)-th future action step.
In our implementation, \(f_{\mathrm{route}}\) is a fixed bounded linear mapping that retains complementary information from both views.

During training, we construct trajectory-derived weak supervision labels 
\(y_t^{(k)} \in \{0,1\}\) from end-effector trajectories, avoiding manual stage annotations. 
Based on the observation that manipulation trajectories alternate between workspace-level motion and local interaction, we initialize all time steps as movement-oriented labels \(0\), and then identify interaction-oriented intervals from end-effector dynamics. 
When gripper-closing signals are available, a short window before closing is treated as a high-confidence interaction interval; otherwise, interaction intervals are detected as low-speed local motions enclosed by higher-speed movements, using translational speed, accumulated path length, and axis-dominance cues. 
Short low-motion pauses that resemble transport pauses are filtered out.
Steps within detected interaction intervals are labeled as \(1\), while the rest remain \(0\). 
These labels are used only as weak supervision, rather than hard semantic stage annotations. 
At inference time, FocusVLA directly predicts continuous intent probabilities from multimodal context and action history, and converts them into step-wise routing weights. Detailed label construction procedures are provided in \hyperref[sec:intent_label_construction]{Appendix~A}.
\subsubsection{Action-Side Dynamic Routing}
Action-Side Dynamic Routing applies the stage-aware control signals produced by the Future Intent Bridge, including the chunk-level intent token and step-wise routing weights, to the action-generation process. Specifically, the action backbone first generates action features over the future horizon conditioned on noisy action tokens~\citep{chi2025diffusion}, the chunk-level intent token, and the diffusion timestep embedding injected through adaptive RMS normalization~\citep{zhang2019root, peebles2023scalable}. These action features are then used as queries to perform cross-attention with global and local visual features, extracting the corresponding global action context \(c_g^{(k)}\) and local action context \(c_l^{(k)}\) for each future action step \(k\).
Unlike methods that directly modify multi-view representations inside the visual encoder, FocusVLA performs view modulation on the action-generation side, allowing visual information to be selected and fused according to the requirements of each action step. Given the step-wise routing weight \(\alpha^{(k)}=[\alpha_g^{(k)},\alpha_l^{(k)}]\) predicted by the Future Intent Bridge, the global and local visual contexts are fused as
\begin{equation}
    c^{(k)} = \alpha_g^{(k)} c_g^{(k)} + \alpha_l^{(k)} c_l^{(k)} .
\end{equation}
The fused context is then injected back into the corresponding action feature through a residual connection and further processed by a lightweight action refinement module for final action prediction.
Through this step-wise routing mechanism, FocusVLA dynamically adjusts the relative contributions of global and local visual information during action generation according to the predicted task progress. This design preserves the stability of the pretrained visual backbone while mitigating the limitation of static fusion, where view dependency is difficult to adapt across different task stages.

\begin{table*}[t]
\centering
\caption{
\textbf{Evaluation on RoboTwin 2.0 simulation tasks.}
We report success rates on all 50 RoboTwin~\citep{chen2025robotwin} tasks, grouped by task horizon. FocusVLA achieves the best overall performance and shows consistent improvements across diverse manipulation tasks.
}
\label{tab:robotwin_horizon_all_50}

\scriptsize
\setlength{\tabcolsep}{1.8pt}
\renewcommand{\arraystretch}{0.62}
\setlength{\aboverulesep}{0.25ex}
\setlength{\belowrulesep}{0.25ex}
\setlength{\extrarowheight}{0pt}

\definecolor{taskgray}{HTML}{E6E6E6}
\definecolor{bestpink}{HTML}{FFC7CE}

\newcommand{\bestcell}[1]{%
  \begingroup
  \setlength{\fboxsep}{0.6pt}%
  \colorbox{bestpink}{\strut\textbf{#1}}%
  \endgroup
}

\newcommand{\taskcell}[2]{\makecell{\textbf{#1}\\[-2pt]\textbf{#2}}}

\resizebox{\textwidth}{!}{%
\begin{tabular}{l*{10}{c}c}
\toprule[2.0pt]

\rowcolor{taskgray}
\multicolumn{12}{c}{\rule{0pt}{1.6ex}\textbf{Short Horizon Tasks}} \\
\midrule
\textbf{Model}
& \taskcell{Click}{Alarmclock}
& \taskcell{Press}{Stapler}
& \taskcell{Turn}{Switch}
& \taskcell{Adjust}{Bottle}
& \taskcell{Grab}{Roller}
& \taskcell{Lift}{Pot}
& \taskcell{Scan}{Object}
& \taskcell{Rotate}{Qrcode}
& \taskcell{Place Object}{Scale}
& \taskcell{Click}{Bell}
& \textbf{Avg.} \\
\midrule
RDT~\citep{liu2025rdt} & 61.0\% & 41.0\% & 35.0\% & 81.0\% & 74.0\% & 72.0\% & 4.0\% & 50.0\% & 1.0\% & \bestcell{80.0\%} & 49.90\% \\
$\pi_0$~\citep{black2024pi_0} & 63.0\% & 62.0\% & 27.0\% & 90.0\% & 96.0\% & 84.0\% & 18.0\% & 68.0\% & 10.0\% & 44.0\% & 56.20\% \\
$\pi_{0.5}$~\citep{intelligence2025pi_} & 28.0\% & 85.0\% & 46.0\% & \bestcell{100.0\%} & \bestcell{100.0\%} & 95.0\% & 35.0\% & \bestcell{82.0\%} & 56.0\% & 4.0\% & 63.10\% \\
\textbf{FocusVLA} & \bestcell{65.0\%} & \bestcell{91.0\%} & \bestcell{53.0\%} & \bestcell{100.0\%} & \bestcell{100.0\%} & \bestcell{98.0\%} & \bestcell{60.0\%} & 66.0\% & \bestcell{71.0\%} & 36.0\% & \bestcell{74.00\%} \\

\midrule
\rowcolor{taskgray}
\multicolumn{12}{c}{\rule{0pt}{1.6ex}\textbf{Medium Horizon Tasks}} \\
\midrule
\textbf{Model}
& \taskcell{Move Can}{Pot}
& \taskcell{Move Pillbottle}{Pad}
& \taskcell{Move Playingcard}{Away}
& \taskcell{Move Stapler}{Pad}
& \taskcell{Place Mouse}{Pad}
& \taskcell{Place A2B}{Left}
& \taskcell{Place A2B}{Right}
& \taskcell{Place Phone}{Stand}
& \taskcell{Place Object}{Stand}
& \taskcell{Place Object}{Basket}
& \textbf{Avg.} \\
\midrule
RDT~\citep{liu2025rdt} & 25.0\% & 8.0\% & 43.0\% & 2.0\% & 1.0\% & 3.0\% & 1.0\% & 15.0\% & 15.0\% & 33.0\% & 14.60\% \\
$\pi_0$~\citep{black2024pi_0} & 58.0\% & 21.0\% & 53.0\% & 0.0\% & 7.0\% & 31.0\% & 27.0\% & 35.0\% & 36.0\% & 16.0\% & 28.40\% \\
$\pi_{0.5}$~\citep{intelligence2025pi_} & 75.0\% & 50.0\% & 83.0\% & 20.0\% & 35.0\% & 62.0\% & 10.0\% & 60.0\% & 77.0\% & 77.0\% & 54.90\% \\
\textbf{FocusVLA} & \bestcell{93.0\%} & \bestcell{79.0\%} & \bestcell{95.0\%} & \bestcell{35.0\%} & \bestcell{47.0\%} & \bestcell{67.0\%} & \bestcell{32.0\%} & \bestcell{63.0\%} & \bestcell{82.0\%} & \bestcell{85.0\%} & \bestcell{67.80\%} \\

\midrule
\rowcolor{taskgray}
\multicolumn{12}{c}{\rule{0pt}{1.6ex}\textbf{Medium-Long Horizon Tasks}} \\
\midrule
\textbf{Model}
& \taskcell{Place Cans}{Plasticbox}
& \taskcell{Place Bread}{Basket}
& \taskcell{Place Bread}{Skillet}
& \taskcell{Place Burger}{Fries}
& \taskcell{Place Container}{Plate}
& \taskcell{Place Can}{Basket}
& \taskcell{Place Empty}{Cup}
& \taskcell{Place}{Fan}
& \taskcell{Place}{Shoe}
& \taskcell{Place Dual}{Shoes}
& \textbf{Avg.} \\
\midrule
RDT~\citep{liu2025rdt} & 19.0\% & 10.0\% & 5.0\% & 50.0\% & 78.0\% & 19.0\% & 56.0\% & 12.0\% & 35.0\% & 4.0\% & 28.80\% \\
$\pi_0$~\citep{black2024pi_0} & 34.0\% & 17.0\% & 23.0\% & 80.0\% & 88.0\% & 41.0\% & 37.0\% & 20.0\% & 28.0\% & 15.0\% & 38.30\% \\
$\pi_{0.5}$~\citep{intelligence2025pi_} & 60.0\% & 74.0\% & 74.0\% & 90.0\% & 95.0\% & 76.0\% & 97.0\% & 44.0\% & 54.0\% & 56.0\% & 72.00\% \\
\textbf{FocusVLA} & \bestcell{90.0\%} & \bestcell{77.0\%} & \bestcell{81.0\%} & \bestcell{95.0\%} & \bestcell{97.0\%} & \bestcell{87.0\%} & \bestcell{99.0\%} & \bestcell{58.0\%} & \bestcell{63.0\%} & \bestcell{60.0\%} & \bestcell{80.70\%} \\

\midrule
\rowcolor{taskgray}
\multicolumn{12}{c}{\rule{0pt}{1.6ex}\textbf{Long Horizon Tasks}} \\
\midrule
\textbf{Model}
& \taskcell{Stack Blocks}{Two}
& \taskcell{Stack Blocks}{Three}
& \taskcell{Stack Bowls}{Two}
& \taskcell{Stack Bowls}{Three}
& \taskcell{Pick Dual}{Bottles}
& \taskcell{Pick Diverse}{Bottles}
& \taskcell{Put Bottles}{Dustbin}
& \taskcell{Blocks Ranking}{Size}
& \taskcell{Blocks Ranking}{Rgb}
& \taskcell{Beat Block}{Hammer}
& \textbf{Avg.} \\
\midrule
RDT~\citep{liu2025rdt} & 21.0\% & 2.0\% & 76.0\% & 51.0\% & 42.0\% & 2.0\% & 21.0\% & 0.0\% & 3.0\% & 77.0\% & 29.50\% \\
$\pi_0$~\citep{black2024pi_0} & 42.0\% & 17.0\% & 91.0\% & 64.0\% & 57.0\% & 27.0\% & 88.0\% & 7.0\% & 19.0\% & 43.0\% & 45.50\% \\
$\pi_{0.5}$~\citep{intelligence2025pi_} & 89.0\% & 60.0\% & 84.0\% & 65.0\% & \bestcell{100.0\%} & 84.0\% & 64.0\% & 42.0\% & 88.0\% & 95.0\% & 77.10\% \\
\textbf{FocusVLA} & \bestcell{99.0\%} & \bestcell{69.0\%} & \bestcell{92.0\%} & \bestcell{74.0\%} & \bestcell{100.0\%} & \bestcell{87.0\%} & \bestcell{96.0\%} & \bestcell{53.0\%} & \bestcell{90.0\%} & \bestcell{96.0\%} & \bestcell{85.60\%} \\

\midrule
\rowcolor{taskgray}
\multicolumn{12}{c}{\rule{0pt}{1.6ex}\textbf{Extra-Long Horizon / Dexterous Tasks}} \\
\midrule
\textbf{Model}
& \taskcell{Shake}{Bottle}
& \taskcell{Shake Bottle}{Horizontally}
& \taskcell{Stamp}{Seal}
& \taskcell{Hanging}{Mug}
& \taskcell{Handover}{Block}
& \taskcell{Handover}{Mic}
& \taskcell{Dump Bin}{Bigbin}
& \taskcell{Open}{Laptop}
& \taskcell{Put Object}{Cabinet}
& \taskcell{Open}{Microwave}
& \textbf{Avg.} \\
\midrule
RDT~\citep{liu2025rdt} & 74.0\% & 84.0\% & 1.0\% & 23.0\% & 45.0\% & 90.0\% & 64.0\% & 59.0\% & 33.0\% & 37.0\% & 51.00\% \\
$\pi_0$~\citep{black2024pi_0} & 97.0\% & 99.0\% & 3.0\% & 11.0\% & 45.0\% & \bestcell{98.0\%} & 83.0\% & 85.0\% & 68.0\% & \bestcell{80.0\%} & 66.90\% \\
$\pi_{0.5}$~\citep{intelligence2025pi_} & 98.0\% & 99.0\% & \bestcell{60.0\%} & 15.0\% & 52.0\% & 48.0\% & 86.0\% & 93.0\% & 55.0\% & 64.0\% & 67.00\% \\
\textbf{FocusVLA} & \bestcell{100.0\%} & \bestcell{100.0\%} & \bestcell{60.0\%} & \bestcell{25.0\%} & \bestcell{63.0\%} & 94.0\% & \bestcell{92.0\%} & \bestcell{98.0\%} & \bestcell{72.0\%} & 61.0\% & \bestcell{76.50\%} \\

\midrule
\textbf{Overall Avg.}
& \multicolumn{2}{c}{\textbf{RDT:} 34.76\%}
& \multicolumn{2}{c}{$\boldsymbol{\pi_0}$\textbf{:} 47.06\%}
& \multicolumn{2}{c}{$\boldsymbol{\pi_{0.5}}$\textbf{:} 66.82\%}
& \multicolumn{3}{c}{\textbf{FocusVLA:} \bestcell{76.92\%}}
& \multicolumn{2}{c}{\bestcell{+10.10\%}} \\
\bottomrule[2.0pt]
\end{tabular}%
}
\vspace{-2.2em}
\end{table*}
\subsection{Training Objective}
FocusVLA is trained with a joint objective for action generation and future-intent prediction. Given the future action target \(U_t\) and intent labels \(Y_t=[y_t^{(1)},\dots,y_t^{(H)}]\), the Future Intent Bridge predicts the intent sequence \(I_t=[i_t^{(1)},\dots,i_t^{(H)}]\), which produces the chunk-level intent token and routing weights \(\alpha_{1:H}\). Conditioned on these signals, the action backbone predicts the denoised action output \(V_t\). We optimize FocusVLA with a joint loss consisting of an action generation loss \(\|V_t-U_t\|_2^2\) and an intent prediction loss \(\lambda_{\mathrm{intent}}\frac{1}{H}\sum_{k=1}^{H}\mathrm{BCE}(i_t^{(k)}, y_t^{(k)})\), where the first term supervises action generation and the second term supervises future-intent prediction with binary labels.
To stabilize training, we first warm up the intent prediction branch for \(S_w\) steps, and then jointly optimize both objectives with occasional teacher forcing to improve training stability.

\label{sec:result}


\section{Experiment}
In this section, we evaluate FocusVLA through simulation, real-world experiments, spatial rearrangement tests, visualization analysis, and ablation studies. 
We aim to answer the following questions:
RQ1: How does FocusVLA compare with existing VLA baselines on large-scale simulation tasks?
RQ2: Does FocusVLA improve real-world manipulation performance?
RQ3: Can FocusVLA adapt to unseen spatial rearrangements?
RQ4: Does FocusVLA dynamically adjust view dependency across task stages?
RQ5: How do its key components contribute to performance?
To ensure a fair comparison, all baseline models and FocusVLA are trained with the same demonstration data, the same number of optimization steps, and evaluated under identical test seeds and evaluation protocols across all tasks and experimental settings, without additional data or tuning.

\subsection{Simulation Evaluation}

We evaluate FocusVLA on the full RoboTwin 2.0 benchmark ~\citep{chen2025robotwin} with 50 manipulation tasks, using 100 trials per task and reporting success rates. Under the same protocol, we compare against RDT, $\pi_0$~\citep{black2024pi_0}, and $\pi_{0.5}$~\citep{intelligence2025pi_}. As shown in Table~\ref{tab:robotwin_horizon_all_50}, FocusVLA achieves the best overall performance, reaching an average success rate of 76.92\% and outperforming the strongest baseline by 10.10 percentage points.
The gains are more pronounced on medium-horizon and long-horizon tasks, suggesting that FocusVLA is especially beneficial in complex manipulation scenarios where the importance of global and local observations changes over time. We further analyze this behavior through visualization and ablation studies in the following sections.

\subsection{Real-World Evaluation}

\begin{figure}[t]
    \centering
    \includegraphics[width=\linewidth]{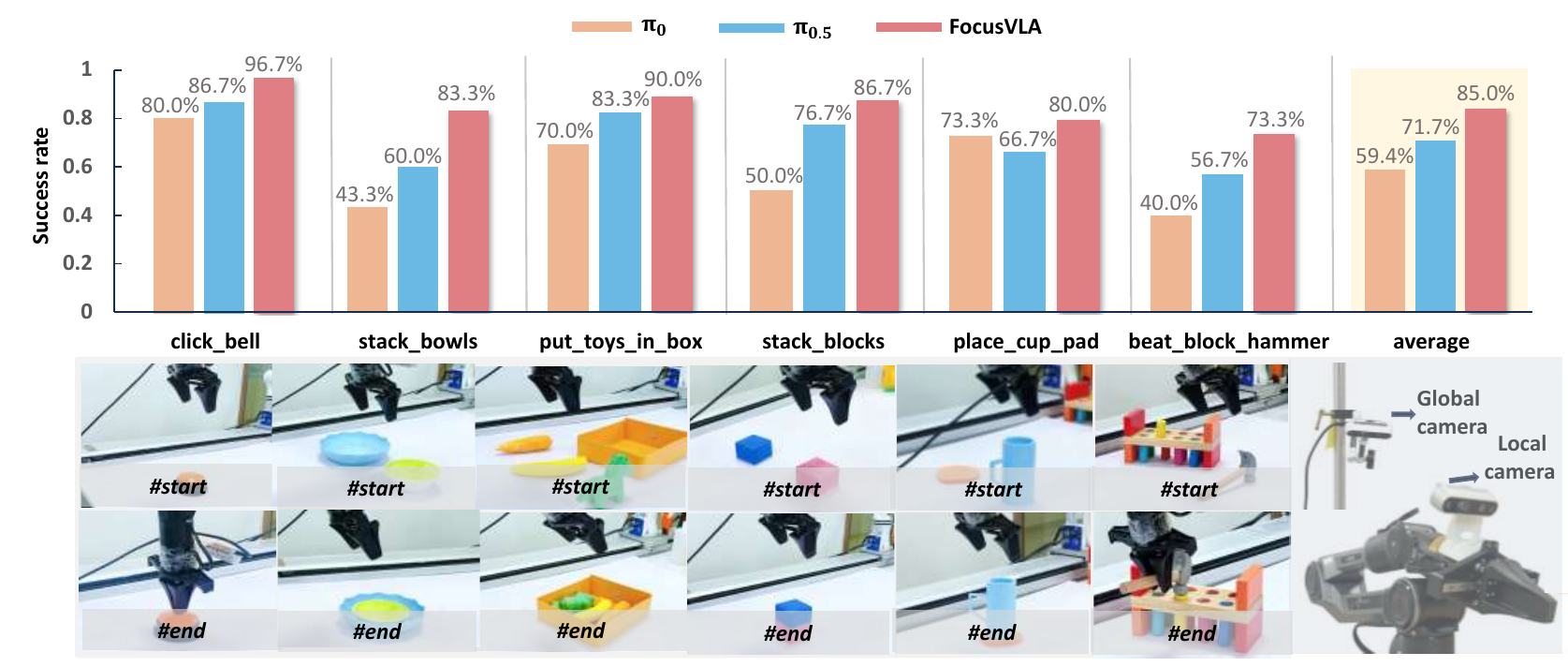}
    \caption{
    \textbf{Real-world evaluation setup and tasks.}
We evaluate FocusVLA on six real-world tasks using a Piper arm with global and local RealSense observations, with 30 trials per task, demonstrating its effectiveness in physical multi-view manipulation.
    }
    \vspace{-0.5em}
    \label{fig:realworld}
\end{figure}

We conduct real-world experiments on a Piper robotic arm equipped with two Intel RealSense D435i cameras for global and local observations. 
As shown in Fig.~\ref{fig:realworld}, we evaluate FocusVLA on six representative manipulation tasks with diverse object configurations and interaction requirements, conducting 30 trials for each task. 
FocusVLA achieves an average success rate of 85.0\%, outperforming $\pi_{0.5}$~\citep{intelligence2025pi_} and $\pi_0$~\citep{black2024pi_0} by 13.3 and 25.6 percentage points, respectively, and obtains the best performance across all six tasks. 
These results suggest that future-intent-guided dynamic view routing improves the robustness of multi-view VLA policies in real-world robot manipulation.

\begin{figure}[t]
    \centering
    \includegraphics[width=0.85\linewidth]{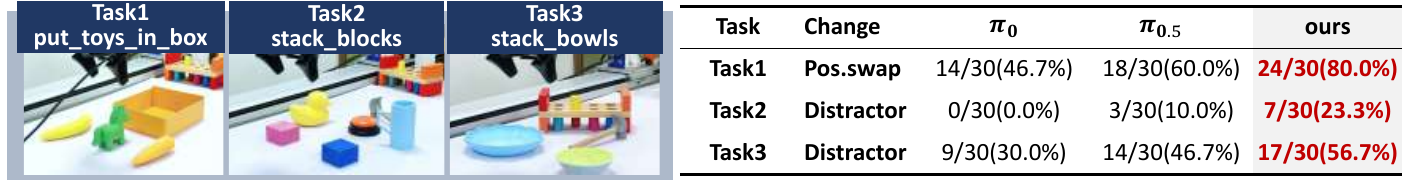}
    \caption{
     Execution adaptability under spatial changes.
    }
    \label{fig:spatial}
    \vspace{-2em}
\end{figure} 

\subsection{Execution Adaptability under Spatial Rearrangement}
As shown in Fig.~\ref{fig:spatial}, we evaluate the execution adaptability of FocusVLA under unseen spatial layouts. Object positions are rearranged, requiring the policy to adapt target localization, approach paths, and fine-grained interactions to the current scene rather than relying on fixed layouts or trajectories from training. Results show that FocusVLA is more stable than static multi-view fusion baselines, suggesting that future action intent modeling and dynamic view routing help the policy better adapt to stage-dependent visual requirements under spatial changes.
\subsection{Visualization Analysis}
\begin{figure}[t]
    \centering
    \includegraphics[width=1\linewidth]{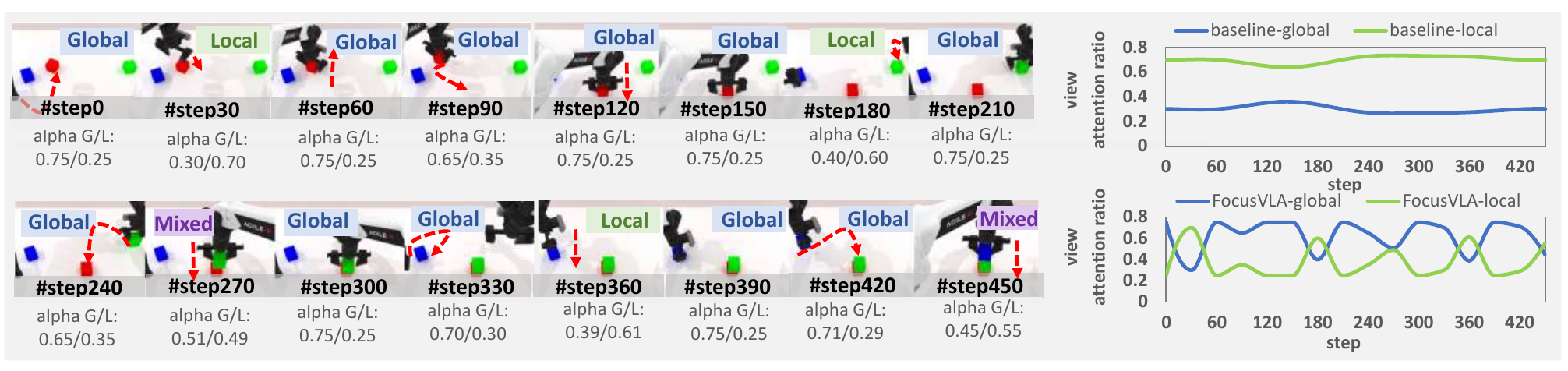}
    \caption{
    \textbf{Visualization of stage-aware view routing.}
We visualize the global/local routing weights of FocusVLA on the \textit{Stack Blocks Three} task. FocusVLA shows adaptive view reallocation across task stages.
}
    \label{fig:visualize}
\end{figure}

Figure~\ref{fig:visualize} visualizes the global/local view routing weights of FocusVLA on the \textit{Stack Blocks Three} task. FocusVLA shows clear stage-aware view reallocation: it relies more on the global view during early target localization and motion planning, and gradually shifts toward the local view during grasping and alignment. During stacking, the routing weights become more balanced, indicating that the model jointly uses global spatial context and local manipulation cues when needed. Overall, the visualization shows that FocusVLA adapts its view usage across task stages rather than maintaining a fixed view bias, consistent with the evolving visual demands of manipulation.
\subsection{Ablation Study}



\begin{table*}[t]
\centering
\caption{Ablations and Future-Intent Contribution Analysis.}
\label{tab:routing_ablation}

\begingroup
\scriptsize
\setlength{\tabcolsep}{3pt}

\begin{tabularx}{0.95\textwidth}{@{}
>{\raggedright\arraybackslash}p{0.11\textwidth}
>{\raggedright\arraybackslash}p{0.15\textwidth}
>{\raggedright\arraybackslash}X
>{\raggedleft\arraybackslash}p{0.19\textwidth}
@{}}
\toprule[1.2pt]

\multicolumn{4}{@{}p{0.95\textwidth}@{}}{%
\centering
\textit{Tasks (2 seeds each; 100 trials/task/seed):}
put\_bottles\_dustbin,
place\_phone\_stand,
move\_can\_pot,
beat\_block\_hammer,
place\_shoe,
open\_laptop,
move\_pillbottle\_pad,
place\_a2b\_left, and
handover\_block.
} \\

\midrule

\textbf{Factor}
& \textbf{Method}
& \textbf{Control}
& \textbf{SR [95\% CI] (\%)} \\

\midrule

Signal
& Learned Static
& Globally shared learned allocation; step-wise application
& 67.5 [65.3, 69.6] \\

Signal
& Reactive-Vision
& Current-vision routing; step-wise application
& 57.4 [55.1, 59.7] \\

Signal
& Time + State
& Time/state routing; step-wise application
& 58.5 [56.2, 60.8] \\

Supervision
& E2E Future
& Action-loss-only routing
& 55.7 [53.4, 58.0] \\

Intent Token
& Routing Only
& Intent-supervised routing without the chunk-level intent token
& 63.3 [61.0, 65.5] \\

Regularizer
& Random Masking
& RDT-style train-only random view dropout
& 54.6 [52.3, 56.9] \\

\textbf{Full}
& \textbf{FocusVLA}
& \textbf{Multimodal future intent; step-wise routing}
& \textcolor{deepred}{\textbf{73.7 [71.6, 75.7]}} \\

\bottomrule[1.2pt]
\end{tabularx}

\endgroup
\end{table*}

To isolate the contribution of FocusVLA, we conduct matched ablations on nine tasks using two training seeds and 100 trials per task per seed; all variants are trained for 20K steps. Table~\ref{tab:routing_ablation} examines the role of FocusVLA’s design choices along four dimensions. First, we replace multimodal future intent with a static view preference, current visual information, or time and state cues to verify the necessity of future intent as the routing signal. Second, we remove explicit future-intent supervision and train the model end-to-end using only the action loss to evaluate the effectiveness of the proposed supervision scheme and rule out the possibility that the performance gains arise merely from increased model capacity. Third, we remove the chunk-level intent token to assess its effectiveness and its complementarity with step-wise routing. Fourth, we replace intent-guided routing with random view dropout to verify the necessity of structured intent guidance for dynamic view modulation and determine whether its benefits can be explained by generic view regularization.

\textbf{Variants.}
Full uses multimodal future-intent supervision, step-wise action-side routing, and a chunk-level intent token. Learned Static retains the global--local fusion structure and intent supervision but replaces sample- and step-dependent routing with a globally shared learnable ratio. Reactive-Vision conditions the step-wise router only on current visual features, while Time+State uses only the execution time and robot state. E2E Future retains the routing architecture and intent-token pathway but removes explicit future-intent supervision, learning the routing variables solely through the action loss. Routing Only removes the chunk-level intent token while retaining intent-supervised step-wise routing. Finally, Random Masking replaces intent-guided routing with RDT-style random view dropout during training.

\textbf{Routing signals and temporal adaptation.}
As shown in Table~\ref{tab:routing_ablation}, Full achieves a mean success rate of 73.7\%, outperforming Learned Static (67.5\%) by 6.2 pp. This gap shows that a globally shared view preference cannot accommodate sample- and step-dependent visual requirements. Full also exceeds Reactive-Vision (57.4\%) and Time+State (58.5\%) by 16.3 and 15.2 pp, respectively. Thus, neither current visual relevance nor time/state shortcuts can substitute for multimodal future intent when determining the visual requirements of future actions.

\textbf{Supervision objective.}
To distinguish the effect of explicit future-intent supervision from that of additional model capacity, E2E Future retains the same routing architecture, step-wise application, and intent-token pathway as Full but is optimized only through the action loss. Its success rate drops to 55.7\%, 18.0 pp below Full, demonstrating that the improvement does not arise merely from introducing an additional routing network; explicit future-intent supervision is essential for learning effective routes.

\textbf{Step- and chunk-level intent conditioning.}
Removing the chunk-level intent token while retaining intent-supervised step-wise routing reduces performance from 73.7\% to 63.3\% (Routing Only), a drop of 10.4 pp. This result shows that the two intent pathways are complementary: step-wise routes determine the view allocation for each future action, while the chunk-level token provides the action generator with an overall estimate of task progress.

\textbf{Comparison with view-dropout regularization.}
RDT-style Random Masking obtains 54.6\% success rate, trailing Full by 19.1 pp. Although random view dropout can reduce reliance on a fixed view during training, it neither predicts which view will be required nor coordinates view allocation across future action steps. Its lower performance therefore indicates that generic view regularization cannot replace intent-guided routing. Overall, these ablations attribute FocusVLA's gains to three complementary components: multimodal prediction of future visual requirements, explicitly supervised step-wise routing, and chunk-level intent conditioning. The improvements cannot be explained by a fixed global--local preference, current-context or time/state shortcuts, additional model capacity, or random view regularization.


	






\section{Conclusion}

\label{sec:conclusion}
We propose \textbf{FocusVLA}, a stage-aware dynamic view routing framework for multi-view VLA policies. By predicting future action intent and converting it into step-wise routing weights, FocusVLA dynamically coordinates global and local visual features during action generation, mitigating the view-dependency inertia caused by static multi-view fusion.
Experiments on RoboTwin and real-world manipulation tasks show that FocusVLA consistently outperforms static multi-view baselines across diverse task horizons and physical scenarios. Visualization results further demonstrate that FocusVLA adapts its view usage across task stages, supporting global planning and local interaction. These results suggest that future-intent-guided routing is an effective strategy for improving the adaptability and robustness of multi-view VLA models.
\section{Limitations}
\label{sec:Limitations}
Although FocusVLA improves stage-aware multi-view utilization, there remain several promising directions for future exploration. 
FocusVLA demonstrates that future-intent-guided routing is an effective and lightweight way to introduce task-progress awareness into multi-view VLA models, opening a practical path toward more adaptive visual reasoning in robotic manipulation. 
Nevertheless, future intent is only one possible representation of manipulation progress. 
We hope this work encourages further exploration of richer progress-aware signals, such as explicit subgoals, object-level interaction states, task-phase descriptions, or language-grounded intermediate plans. 
Extending such progress-aware routing mechanisms to cross-embodiment settings with diverse robots, camera configurations, and real-world environments is also an important direction for future work.

\clearpage
\acknowledgments{
This work was supported by the Science and Technology Major Project of Jiangsu Province (Grant No. BG2024041). We gratefully acknowledge this support, which facilitated the research, system development, and experimental validation presented in this work. We sincerely thank the Area Chair and anonymous reviewers for their careful reading, thoughtful questions, and constructive suggestions, which helped us clarify the motivation, strengthen the empirical evaluation, and improve the presentation of this work. 
}


\bibliography{example}  

\clearpage
\begin{center}
    \Large\bfseries Appendix
\end{center}
\vspace{1em}
\section*{A. Automatic Intent Label Construction}
\label{sec:intent_label_construction}
\begin{algorithm}[h]
\caption{Automatic Intent Label Construction}
\label{alg:auto_intent_label}
\begin{algorithmic}[1]
\FOR{each episode}
    \STATE compute smoothed translational speed and rotation for both arms
    \STATE combine the two arms by taking the maximum speed/rotation at each frame

    \STATE detect gripper-closing windows if gripper signals are available
    \IF{any valid closing window exists}
        \STATE expand each closing run by a 16-frame lookback
        \STATE merge same-arm windows with gap $\leq 6$
        \STATE use the resulting windows as interaction seeds
    \ELSE
        \STATE compute high-speed threshold as the 70th percentile of speed
        \STATE compute low-speed threshold as the 45th percentile of speed
        \STATE extract high-speed segments of length $\geq 3$
        \STATE extract low-speed segments of length $\geq 6$
        \STATE keep only low-speed segments that are surrounded by high-speed motion
        \STATE reject low-speed segments whose path length exceeds the adaptive cap unless they are strongly axis-dominant
        \STATE add press-like candidates that occur late, follow an approach, and have small path length
        \STATE add handover-like windows based on end-effector proximity and cooperative motion
    \ENDIF

    \STATE merge nearby interaction windows with gap tolerance 12
    \STATE split each merged interaction by high-speed subsegments
    \STATE demote non-dominant high-speed chunks to global movement
    \STATE remove short interaction valleys that are better explained as transport pauses
    \STATE trim interaction windows around valid gripper-closing events when available

    \STATE assign label 1 to all frames inside final interaction windows
    \STATE assign label 0 to all remaining frames
    \STATE save the binary label sequence
\ENDFOR
\end{algorithmic}
\end{algorithm}
We automatically construct binary intent labels from demonstration trajectories without manual phase annotation. The label generation pipeline first applies an end-pose segmentation backend to each episode and then converts the resulting temporal segments into frame-wise supervision. The segmentation operates on the end-effector motion of both arms and uses a set of deterministic heuristics to distinguish \emph{global movement} from \emph{interaction}.

Let $x_t^a \in \mathbb{R}^3$ and $q_t^a \in \mathbb{R}^4$ denote the translation and quaternion orientation of arm $a \in \{\mathrm{left}, \mathrm{right}\}$ at time step $t$. We compute the per-arm translational speed and rotation increment as
\begin{equation}
v_t^a = \left\|x_t^a - x_{t-1}^a\right\|_2,
\qquad
r_t^a = 2\arccos\!\left(\left|\langle q_{t-1}^a, q_t^a\rangle\right|\right),
\end{equation}
and then smooth both signals using a 5-tap moving average. The fused motion signal is obtained by taking the maximum over the two arms:
\begin{equation}
v_t = \max\!\left(v_t^{\mathrm{left}}, v_t^{\mathrm{right}}\right),
\qquad
r_t = \max\!\left(r_t^{\mathrm{left}}, r_t^{\mathrm{right}}\right).
\end{equation}

\paragraph{Gripper-driven interaction seeds.}
When gripper signals are available, interaction detection is dominated by closing events. We detect a closing event when the per-step gripper difference satisfies
\begin{equation}
\Delta g_t < -0.03
\end{equation}
for at least 4 consecutive frames. Closures occurring before the first 12 frames are ignored as initialization noise. Each valid closing run is expanded by a 16-frame lookback window, and nearby windows of the same arm are merged if their temporal gap is at most 6 frames. If any gripper-closing window exists, these windows are directly treated as interaction seeds and the remaining timeline is assigned to global movement unless further trimming is needed.

\paragraph{Motion-driven interaction seeds.}
If no valid gripper-close evidence is available, we fall back to motion-based segmentation. We define high-speed frames using the 70th percentile of the smoothed speed signal and low-speed frames using the 45th percentile:
\begin{equation}
\text{high-speed}:\ v_t \ge p_{70}(v), 
\qquad
\text{low-speed}:\ v_t \le p_{45}(v).
\end{equation}
High-speed segments must last at least 3 frames, while low-speed segments must last at least 6 frames. A low-speed segment $[s,e]$ is considered an interaction candidate only if it is temporally surrounded by high-speed motion, i.e., there exists high-speed activity within a 20-frame lookback before $s$ and within a 20-frame lookahead after $e$.

For each low-speed candidate, we compute its fused path length
\begin{equation}
P(s,e) = \sum_{t=s}^{e}\max\!\left(\|x_t^{\mathrm{left}}-x_{t-1}^{\mathrm{left}}\|_2,\ \|x_t^{\mathrm{right}}-x_{t-1}^{\mathrm{right}}\|_2\right),
\end{equation}
and compare it against an adaptive path cap
\begin{equation}
P_{\max} = \max\!\left(0.02,\ 2\cdot p_{80}\big(\{P(s,e)\}_{\text{low-speed segments}}\big)\right).
\end{equation}
A low-speed segment is rejected if $P(s,e) > P_{\max}$ and its dominant-axis ratio is below 0.78. The dominant-axis ratio is defined as the fraction of motion energy along the most active axis within the segment. Intuitively, this rule preserves compact, axis-consistent local adjustments while filtering out long translational motions.

\paragraph{Press-like and handover-like cases.}
Some tasks do not exhibit a clear gripper-closing event but still contain local interactions such as pressing or handover. If no interaction candidate is found and no closing window exists, we additionally keep a late low-speed segment as a press-like interaction if it satisfies all of the following conditions: (i) its start index is at least 12, (ii) it is preceded by a high-speed approach within a 30-frame lookback, and (iii) its fused path length is smaller than $\max(0.012, 0.8P_{\max})$. We further allow a short low-speed hold tail after such a press-like interaction when the following low-speed segment starts within 6 frames and its length is no more than 14 frames.

We also add handover-style windows based on near-field dual-arm coordination. These windows are detected when the distance between the two end-effectors becomes sufficiently small and the two arms move cooperatively. A stricter handover detector uses an arm-distance threshold clipped to $[0.16, 0.25]$, requires a minimum overlap of motion share, and expands each candidate window by 6 frames. A looser fallback detector uses a distance threshold clipped to $[0.22, 0.30]$ and retains only windows whose motion share, activity mean, and activity peak exceed adaptive thresholds. Nearby handover windows are merged if their gap is within 8 frames.

\paragraph{Interaction cleanup and priority rules.}
After collecting all interaction seeds, we merge nearby or overlapping interaction windows with a 12-frame gap tolerance. To prevent large translations from being mislabeled as interaction, each merged interaction window is further split by high-speed subsegments using a movement threshold at the 72nd percentile of $v_t$. High-speed subsegments shorter than 7 frames are ignored, and interaction fragments shorter than 6 frames are discarded. Within such high-speed subsegments, we keep them as interaction only if their dominant-axis ratio is at least 0.78; otherwise, they are demoted to global movement.

We then apply three additional reclassification rules to remove spurious short pauses. First, a short interaction valley between two strong global segments is converted to global if its duration is below 16 frames and its fused path length is sufficiently small. Second, if an interaction valley is short (at most 18 frames), low-path, and its neighboring global segments move along the same axis and in the same direction, it is also reclassified as global. Third, a short low-motion valley between two strong global transport segments is demoted to global if its length is at most 12 frames and its path length is below 0.03. Finally, when gripper-closing evidence exists, interaction windows are trimmed around the closing intervals so that only the frames consistent with the closing event are preserved.

Given the final interaction intervals $\mathcal{I} = \{[s_k,e_k]\}_{k=1}^{K}$, we define the binary intent label sequence $\mathbf{y}\in\{0,1\}^T$ as
\begin{equation}
y_t =
\begin{cases}
1, & \text{if } t \in \bigcup_{k=1}^{K}[s_k,e_k], \\
0, & \text{otherwise},
\end{cases}
\qquad t=1,\dots,T.
\end{equation}
Here, $y_t=1$ denotes local interaction and $y_t=0$ denotes global movement. The same deterministic procedure is applied to every episode, making the generated labels fully reproducible.

\section*{B. Training and Evaluation Configurations}
\label{sec:training_eval_configs}

We train FocusVLA separately on the RoboTwin simulated benchmark and the real-world benchmark. Within each benchmark, we collect 50 demonstration trajectories per task and jointly train a single policy across all tasks in that benchmark. In other words, simulation tasks are trained in one multi-task setting, while real-world tasks are trained in another, with no data mixing across the two domains. Training is conducted on 8 NVIDIA A100 80GB GPUs for 100{,}000 optimization steps.

To stabilize optimization, we adopt an intent warmup strategy. Specifically, during the initial warmup phase, we optimize the future intent branch first so that the model can learn a reliable interaction-pattern prior before being used to guide action generation. After warmup, we jointly optimize the action-generation objective and the intent-prediction objective.

Since interaction frames are typically less frequent than global movement frames, we further apply a class-balanced intent loss to mitigate label imbalance. Let $p_t$ denote the predicted intent probability at time step $t$ and $y_t \in \{0,1\}$ denote the corresponding binary label, where $y_t=1$ indicates interaction and $y_t=0$ indicates global movement. The intent loss is formulated as a weighted binary cross-entropy:
\begin{equation}
\mathcal{L}_{\text{intent}} = - \frac{1}{T} \sum_{t=1}^{T}
\left(
\lambda_{+} y_t \log p_t
+
\lambda_{-} (1-y_t)\log(1-p_t)
\right),
\end{equation}
where $\lambda_{+}$ and $\lambda_{-}$ are class weights for interaction and movement, respectively. In practice, these weights are set to balance the contribution of the two classes and prevent the model from being dominated by the majority movement labels.

The overall training objective is
\begin{equation}
\mathcal{L}_{\text{total}} =
\mathcal{L}_{\text{act}} + \lambda_{\text{intent}} \mathcal{L}_{\text{intent}},
\end{equation}
where $\mathcal{L}_{\text{act}}$ denotes the action-generation loss and $\lambda_{\text{intent}}$ controls the relative importance of the intent supervision.

During evaluation, we report the success rate on the corresponding task-specific test episodes using the checkpoint obtained from the respective domain-specific training. Unless otherwise stated, the evaluation protocol follows the standard benchmark setting described in the main paper.

\section*{C. Future Intent Bridge Details}
\label{sec:future_intent_bridge_details}

The Future Intent Bridge is designed to infer how the interaction mode evolves over the future action horizon from the current multimodal context. Given the contextualized language features from the VLA prefix backbone, together with visual features and state-history features, we use language tokens as queries and visual/state-history tokens as keys and values in a cross-modal attention layer. This design allows the bridge to selectively retrieve scene evidence and recent motion cues that are most relevant to the instructed task.

The resulting multimodal representation is then fed into a temporal intent head, which predicts a step-wise intent sequence over the future horizon,
\begin{equation}
    I_t = [i_t^{(1)}, i_t^{(2)}, \dots, i_t^{(H)}],
\end{equation}
where $i_t^{(k)} \in [0,1]$ denotes the predicted probability that the $k$-th future action step requires local interaction-oriented visual evidence. Unlike a discrete stage label, this sequence provides a continuous estimate of how interaction demand evolves across the future horizon.

To convert the intent sequence into downstream control signals, we further derive two outputs from it. First, we compress the future intent sequence into a chunk-level intent token, which serves as a coarse semantic condition for the action suffix. Second, we map the predicted intent probabilities to step-wise routing weights,
\begin{equation}
    \alpha^{(k)} = [\alpha_g^{(k)}, \alpha_l^{(k)}],
\end{equation}
where $\alpha_g^{(k)}$ and $\alpha_l^{(k)}$ denote the relative emphasis on global and local visual evidence for the $k$-th future step, respectively. These routing weights are obtained by linearly mapping the predicted intent probabilities into the allowed global/local range and are subsequently used by the action-side routing module.

\section*{D. Action-Side Dynamic Routing Details}
\label{sec:action_side_dynamic_routing_details}

Action-Side Dynamic Routing applies the stage-aware control signals produced by the Future Intent Bridge to the action generation process. After the action backbone processes the noisy action tokens under the conditioning of the chunk-level intent token and the diffusion timestep embedding, it produces action features for the future horizon. These action features are then used as queries to attend to global and local visual features separately, yielding two context streams for each future step:
\begin{equation}
    c_g^{(k)}, \qquad c_l^{(k)}, \qquad k=1,\dots,H.
\end{equation}

Different from static multi-view fusion, we perform view modulation on the action side, so that the model can decide at each future step which visual source should contribute more strongly. Given the routing weights $\alpha^{(k)} = [\alpha_g^{(k)}, \alpha_l^{(k)}]$, the two visual contexts are fused as
\begin{equation}
    c^{(k)} = \alpha_g^{(k)} c_g^{(k)} + \alpha_l^{(k)} c_l^{(k)}.
\end{equation}
The fused context is then injected back into the corresponding action feature through a gated residual connection, producing an enhanced action representation for final prediction. This design allows FocusVLA to dynamically shift its dependence between global and local views across different manipulation stages, while preserving the pretrained visual backbone and avoiding rigid, hand-designed view selection rules.

\section*{E. BFA-Related Controls}

We further conduct matched controls to distinguish FocusVLA from BFA along four dimensions: prediction horizon, fusion site, label construction, and intent conditioning. Within this comparison, all variants use the same four-task subset and evaluation protocol.

\textbf{Prediction horizon.}
\begin{table*}[t]
\centering
\caption{BFA-related controls.}
\label{tab:controlled_ablation}

\begingroup
\scriptsize
\setlength{\tabcolsep}{3pt}

\begin{tabular}{@{}
>{\raggedright\arraybackslash}p{0.18\textwidth}
>{\raggedright\arraybackslash}p{0.13\textwidth}
>{\raggedright\arraybackslash}p{0.11\textwidth}
>{\centering\arraybackslash}p{0.06\textwidth}
>{\raggedleft\arraybackslash}p{0.17\textwidth}
@{}}
\toprule[1pt]

\multicolumn{5}{@{}p{0.70\textwidth}@{}}{%
\centering
\textit{Tasks (100 trials/task):}
open\_microwave,
pick\_diverse\_bottles,
stack\_bowls\_two, and
pick\_dual\_bottles.
} \\

\midrule

\textbf{Variant}
& \textbf{Routing signal}
& \textbf{Application}
& \textbf{Token}
& \textbf{SR [95\% CI] (\%)} \\

\midrule

Reactive-Vision
& Current vision
& Step-wise
& Yes
& 61.3 [56.4, 65.9] \\

Reactive (\(H_r{=}1\))
& MM intent
& Shared
& Yes
& 55.5 [50.6, 60.3] \\

Feature-side
& MM intent
& Pre-policy
& Yes
& 67.0 [62.2, 71.4] \\

BFA-style Labels
& BFA-style
& Step-wise
& Yes
& 72.0 [67.4, 76.2] \\

Reactive w/o Token
& MM intent
& Shared
& No
& 49.8 [44.9, 54.6] \\

Routing Only
& MM intent
& Step-wise
& No
& 71.5 [66.9, 75.7] \\

\textbf{Full}
& \textbf{MM intent}
& \textbf{Step-wise}
& \textbf{Yes}
& \textcolor{deepred}{\textbf{79.5 [75.3, 83.2]}} \\

\bottomrule[1pt]
\end{tabular}

\endgroup
\end{table*}
BFA estimates one view allocation from current vision and applies the resulting fused representation to the entire action chunk. It therefore cannot condition routing on instruction-, state-, or history-dependent requirements or explicitly adapt the allocation across future steps within the chunk. Reactive-Vision retains step-wise routing but uses only current vision, whereas Reactive (\(H_r{=}1\)) retains multimodal intent conditioning but predicts one route shared by the full \(H\)-step chunk. Full outperforms these variants by 18.2 and 24.0 pp, respectively, supporting the contributions of multimodal future-intent conditioning and anticipatory step-specific routing.

\textbf{Fusion site.}
BFA merges multi-view features before the policy, whereas FocusVLA maintains separate global and local action contexts and fuses them for each predicted action step. With routing, intent supervision, and the chunk-level token fixed, Feature-side moves only the fusion operation to the pre-policy location used by BFA. Its 12.5-pp gap from Full isolates the benefit of performing view fusion within action generation.

\textbf{Label construction.}
BFA uses offline VLM annotations of robot states together with task-specific rule mappings. Because BFA provides neither usable code nor complete labeling rules, we implement a paper-based BFA-style baseline using GPT-4o arm-state annotations and task-specific mappings. With step-wise routing and the intent token fixed, Full exceeds BFA-style Labels by 7.5 pp. The human-agreement evaluation in Table~\ref{tab:intent_robustness} further supports the accuracy of our task-agnostic trajectory labels.

\textbf{Intent token.}
Unlike BFA, FocusVLA provides the action generator with a chunk-level intent token in addition to step-wise routing weights. Adding this token improves Reactive (\(H_r{=}1\)) over Reactive w/o Token by 5.7 pp and Full over Routing Only by 8.0 pp. These matched comparisons show that chunk-level intent conditioning benefits both chunk-shared and step-wise routing.

Together, these controls show that anticipatory routing, action-side fusion,
trajectory-derived supervision, and chunk-level intent conditioning each
provide distinct gains.

\section*{F. Intent-Label Evaluation}

We evaluate the trajectory-derived intent labels from three perspectives:
agreement with human annotations, held-out intent-prediction accuracy, and
robustness to label definitions and prediction errors. The robustness
experiments use the same four tasks as the BFA-related controls, with
50 trials per task.

\textbf{Human agreement.}
Using human annotations as reference, our task-agnostic trajectory labels
achieve 82.9\% F1, exceeding the paper-based BFA-style labels by 17.8
points. This result indicates that our labels better capture stage-specific
visual requirements without relying on offline VLM annotations or
task-specific rule mappings.

\textbf{Intent-prediction accuracy.}
On held-out trajectories, the Future Intent Bridge achieves 78.4\% F1 at
a 0.5 classification threshold and 91.2\% AUROC. Qualitative inspection
shows that most prediction errors slightly shift transition boundaries
rather than confuse the underlying interaction stages, suggesting that
the predicted scores preserve the temporal structure required for
step-wise routing.

\textbf{Robustness to label definitions.}
We construct loose and strict variants by relaxing or tightening the
local-interaction criteria used for trajectory labeling. Relative to the
default definition, these variants alter 45.2\% and 27.6\% of the labels,
respectively. Nevertheless, their success rates are 75.5\% and 74.0\%,
compared with 78.0\% under the default labels, corresponding to a maximum
change of 4.0 pp. Thus, downstream control performance is relatively
insensitive to the precise heuristic thresholds used to construct the
intent labels.

\textbf{Robustness to prediction errors.}
To evaluate whether FocusVLA depends on perfectly accurate intent estimates,
we independently corrupt 10\% and 20\% of the predicted intents at inference.
The corresponding success rates are 73.5\% and 75.5\%, differing from the
uncorrupted result by at most 4.5 pp. FocusVLA therefore remains effective
under moderate intent-prediction errors.

Overall, our trajectory-derived supervision aligns more closely with human
annotations than the BFA-style alternative, supports accurate held-out intent
prediction, and remains robust to both label-definition choices and moderate
prediction errors. Generalization to unseen camera layouts remains future work.
\begin{table*}[t]
\centering
\caption{Intent-label evaluation.}
\label{tab:intent_robustness}

\begingroup
\scriptsize
\setlength{\tabcolsep}{3pt}

\begin{tabular*}{0.6\columnwidth}
{@{\extracolsep{\fill}}lccc@{}}
\toprule[1pt]

\textbf{Metric}
& \textbf{BFA-style}
& \textbf{Ours}
& \textbf{Effect} \\

\midrule

Human F1 \(\uparrow\) (\%)
& 65.1
& \textcolor{deepred}{\textbf{82.9}}
& \textbf{+17.8} \\

Intent F1/AUROC (\%)
& N/A
& 78.4/91.2
& N/A \\

Thresh. SR (D/L/S) (\%)
& N/A
& 78.0/75.5/74.0
& \(\leq 4.0\) pp \\

Infer. SR (0/10/20\%) (\%)
& N/A
& 78.0/73.5/75.5
& \(\leq 4.5\) pp \\

\bottomrule[1pt]
\end{tabular*}

\endgroup
\end{table*}

\section*{G. Architecture and Implementation Details}

\textbf{Architecture.}
FocusVLA augments \(\pi_{0.5}\) with a Future Intent Bridge and an
action-side router. In the Bridge, the Temporal State-History Encoder
linearly embeds the current robot state into a 2,048-D state token.
It separately processes the previous 30 actions and their temporal
differences using two local temporal blocks, followed by softmax pooling
to produce a 2,048-D history token. Together with the multimodal
vision-language context, these tokens are processed by four-head,
2,048-D cross-attention. Masked-mean pooling aggregates the valid
features, and a \(2048\!\rightarrow\!4096\!\rightarrow\!30\) MLP predicts
intent scores for \(H=30\) future action steps. The action-side router
uses four-head, 1,024-D cross-attention to construct separate global
and local action contexts, which are fused for each predicted action
step using the corresponding routing weight. The Bridge additionally
provides a chunk-level intent token to condition the overall action
generation process.

\textbf{Training and computational overhead.}
The intent objective uses a loss weight of
\(\lambda_{\mathrm{intent}}=1\) and a positive-class weight of
\(w_{+}=15\). Training begins with a 5K-step intent-only warm-up.
From steps 5K to 15K, the teacher-forcing probability is annealed from
0.77 to 0.3 and then remains at 0.3 for the rest of training; teacher
forcing is disabled at inference. FocusVLA adds 70.13M parameters to
\(\pi_{0.5}\). At batch size \(B=1\) on one NVIDIA A100, its per-chunk
latency is 92.78 ms, compared with 79.59 ms for \(\pi_{0.5}\), corresponding
to an increase of 13.19 ms (16.6\%).

\textbf{Bounded routing weights.}
For each sample and future step, the Bridge predicts an intent logit
from multimodal context. A sigmoid converts this logit into a probability,
which is linearly mapped to a view weight strictly within
\((0.25,0.75)\); the complementary view receives the remaining weight,
so the two weights sum to one. This task-invariant mapping can favor
one view by up to a \(3{:}1\) ratio while retaining at least 25\% of the
complementary view, thereby preserving complementary evidence and
avoiding brittle \(0/1\) selection. The bounds and scalar mapping are
shared across tasks; context dependence arises from the sample- and
step-specific intent predictions rather than from task-specific
endpoints. At the reporting precision, 64.19\% of held-out
\(\alpha_g\) values remain strictly interior, while the remaining values
round to 0.25 or 0.75. These endpoint occurrences therefore represent
confident, near-binary intent predictions rather than a fixed routing
policy.

\end{document}